\documentclass[]{interact}

\usepackage{float}
\usepackage{enumitem}
\usepackage{graphicx}
\usepackage{booktabs}  
\usepackage{siunitx}   

\usepackage{epstopdf}
\usepackage[caption=false]{subfig}

\usepackage[numbers,sort&compress]{natbib}
\bibpunct[, ]{[}{]}{,}{n}{,}{,}
\renewcommand\bibfont{\fontsize{10}{12}\selectfont}

\theoremstyle{plain}

\theoremstyle{definition}

\theoremstyle{remark}

\usepackage{tikz}

\begin{document}

\articletype{ARTICLE TEMPLATE}

\title{fakenewsbr: A Fake News Detection Platform for Brazilian Portuguese}

\author{
\name{Luiz Giordani\textsuperscript{a};  Gilsiley Darú\textsuperscript{a};  Rhenan Queiroz\textsuperscript{a}; Vitor Buzinaro\textsuperscript{a}; Davi Keglevich Neiva\textsuperscript{a}; Daniel Camilo Fuentes Guzm\'{a}n\textsuperscript{a}\textsuperscript{b}\thanks{CONTACT Daniel Camilo Fuentes Guzm\'{a}n. Email: camiguz89@usp.br}; Marcos Jardel Henriques\textsuperscript{a,b}; Oilson Alberto Gonzatto Junior\textsuperscript{a} and Francisco Louzada\textsuperscript{a}}
\affil{\textsuperscript{a}Department of Applied Mathematics and Statistics, Institute of Mathematical and Computer Sciences, University of São Paulo, São Carlos, Brazil; \textsuperscript{b}Department of Statistics, Federal University of São Carlos, S\~ao Paulo, Brazil.}
}

\maketitle

\begin{abstract}
The proliferation of fake news has become a significant concern in recent times due to its potential to spread misinformation and manipulate public opinion. This paper presents a comprehensive study on detecting fake news in Brazilian Portuguese, focusing on journalistic-type news. We propose a machine learning-based approach that leverages natural language processing techniques, including TF-IDF and Word2Vec, to extract features from textual data. We evaluate the performance of various classification algorithms, such as logistic regression, support vector machine, random forest, AdaBoost, and LightGBM, on a dataset containing both true and fake news articles. The proposed approach achieves high accuracy and F1-Score, demonstrating its effectiveness in identifying fake news.
Additionally, we developed a user-friendly web platform, fakenewsbr.com, to facilitate the verification of news articles' veracity. Our platform provides real-time analysis, allowing users to assess the likelihood of fake news articles. Through empirical analysis and comparative studies, we demonstrate the potential of our approach to contribute to the fight against the spread of fake news and promote more informed media consumption.

\end{abstract}

\begin{keywords}
Fake News; Artificial intelligence; Machine Learning; Natural Language Processing; Neural Networks; Classification Algorithms.
\end{keywords}

\section{Introduction}
In an era characterized by the rapid dissemination of information through online platforms, the challenge of distinguishing between accurate and false information has become increasingly prominent. Fake news, defined as deliberately fabricated information presented as genuine news, can potentially deceive and mislead the public, often resulting in detrimental consequences for society. The widespread impact of fake news on public opinion, political discourse, and social dynamics necessitates effective strategies for its detection and mitigation \cite{kim2021systematic, raza2022fake, gomes2020scientific}.

The proliferation of fake news has gained significant attention in the context of various global events, such as the COVID-19 pandemic, geopolitical conflicts, and elections \cite{baarir2021fake}. These scenarios highlight the urgency of addressing the spread of misinformation and disinformation. Consequently, researchers and practitioners have turned to artificial intelligence and machine learning techniques to develop robust tools for identifying and combating fake news.

In the second half of 2021, amid the global COVID-19 pandemic, a collaboration between researchers from the Institute of Mathematical and Computer Sciences at the University of São Paulo (ICMC-USP) and students enrolled in the Professional Master's Degree in Mathematics, Statistics and Computing Applied to Industry (MECAI-USP) program gave rise to a pivotal objective: to combat the escalating dissemination of fake news \cite{monteiro2018contributions}. This tumultuous period gave rise to the need to develop a platform designed to detect fake news\cite{haridas2019detecting, silva2020analise}. This article elucidates the journey of creating such a platform, executed through the adoption of agile project management methodologies\cite{rasnacis2017method, moloto2020impact} and the implementation of the Cross-Industry Standard Process for Data Mining (CRISP-DM) framework\cite{8943998}.

This paper presents a comprehensive study on detecting fake news in Brazilian Portuguese, focusing on journalistic-type news articles. We propose a machine learning-based approach that combines natural language processing techniques with classification algorithms. Our approach aims to provide an efficient and accurate means of identifying fake news, thereby contributing to the broader goal of promoting informed media consumption and countering the harmful effects of misinformation.

\subsection{Related Work}
The detection of fake news has garnered significant interest within the research community. Various methodologies have been explored to address this challenge, encompassing linguistic analysis, network-based approaches, and machine learning techniques \cite{silva2020towards, monteiro2018contributions}. Linguistic analysis identifies linguistic patterns and features that distinguish fake news from legitimate content \cite{abonizio2020language}. Network-based approaches analyze information propagation and diffusion patterns across social networks to identify sources of misinformation \cite{santos2020fact}. However, these methods may have limitations in handling the complex and dynamic nature of fake news dissemination \cite{figueira2017current}.

Machine learning techniques, mainly supervised classification, have shown promise in detecting fake news \cite{faustini2020fake, lilleberg2015support}. These approaches leverage labelled datasets to train models that can generalize patterns and characteristics of fake news articles. Existing research has explored using different features, such as textual content, metadata, and user engagement metrics, to develop effective classification models.
Our work builds upon and extends prior research by focusing on the specific context of Brazilian Portuguese journalistic-type news articles. We employ a combination of natural language processing techniques and machine learning algorithms to extract meaningful features from textual data, allowing us to distinguish between true and fake news articles with high accuracy.


This article is organized as follows: In Section \ref{MaterialandMethods}, we describe the dataset used, text preprocessing, web scraping, Word2Vec, and classification models developed for this project. Section \ref{resultsandDiscussion} presents the results obtained, including performance metrics for the various models tested. In Section \ref{PlatformforAutomaticIdentificationofFakeNews}, we provide a detailed description of the platform developed for automatic fake news detection. In Section \ref{ConclusionandFutureWork}, we summarize the conclusions based on the results and outline potential future research directions. Appendix A, titled "Structure of the Fake News Detection Model," provides additional information about the structure of the developed detection models. Appendix B, titled "Platform for Automatic Identification of Fake News," describes the platform and provides a step-by-step guide on its usage.





\section{Material and Methods} \label{MaterialandMethods}

\subsection{Dataset}
For our study, we utilized the dataset constructed by \cite{silva2020towards}, which consists of $7,200$ textual news articles in Brazilian Portuguese. These articles were extracted through web scraping from sources available on the internet, spanning the years $2016$ to $2018$. The dataset comprises an equal number of true and fake news articles ($3,600$ each). To ensure balanced and relevant content, the authors manually analyzed and selected news articles, ensuring each true news article has a corresponding fake news article with similar themes and lengths. Partially false news articles were excluded from the dataset to maintain a clear distinction between true and fake news.

To incorporate more recent information and vocabulary, we performed web scraping from websites associated with fake news and those identified as reputable by fact-checking agencies. Such additional data collection aimed to enrich our corpus with up-to-date vocabulary and contextual relationships.


\subsection{Text pre-processing}
Text preprocessing is a crucial step in preparing the textual data for analysis \cite{preprocessing2015vijayarani}. We followed common preprocessing steps, including tokenization, stop word removal, word stemming, and applying TF-IDF algorithms \cite{bird2009natural}.
Tokenization involves extracting individual words, tags, and apostrophes from the text, allowing subsequent analysis. Stop word removal eliminates common words that do not contribute significant meaning, using the Natural Language Toolkit (NLTK) library for Portuguese. Word stemming reduces words to their root forms, aiding in identifying word variations.

\subsection{Web Scraping}
We conducted web scraping (\cite{zhao2017web}) to collect news articles from various sources, including mainstream, credible, and dubious websites. Our sources were selected based on research conducted by independent fact-checking organizations. The collected dataset comprised approximately $99,123$ news articles covering the period from $2017$ to $2021$, representing a diverse range of topics and sources.

Extracting data from the \emph{World Wide Web} and storing it in files or databases is a technique called \emph{Web Scraping} in the literature, as presented by \cite{zhao2017web}. In the context of this article, the purpose of using web scraping is to build a database of news articles from mainstream, credulous, and doubtful sources. The doubtful sources were chosen according to research from the independent media outlet and fact-checking website \emph{Aos fatos}.


\subsection{Word2Vec}
We employed Word2Vec, a neural network architecture, to map words from our corpus into vector representations based on their contextual relationships \cite{mikolov2013efficient}, \cite{gao2019context}. This technique captures semantic meanings and relationships between words, enabling us to measure word similarity based on vector proximity \cite{rehurek2011gensim}. Two Word2Vec architectures, Continuous Bag of Words (CBOW) and Skip-Gram, were used to create word embeddings for feature extraction. The Word2Vec models are trained on the corpus assembled from the web scraping process. Figure \ref{fig:figura-lgbm_w2v} shows the training of the model.



\subsection{Classification Models}
We evaluated the performance of various classification models on our dataset, including logistic regression, support vector machine (SVM), random forest, AdaBoost, and LightGBM. The features for the models were constructed using both TF-IDF and Word2Vec representations. We conducted a 5-fold cross-validation to assess model performance and used metrics such as accuracy, precision, recall, and F1-Score.

\section{Results and Discussion} \label{resultsandDiscussion}

The table (\ref{tab:comparacaotodosmodelos}) showcases the outcomes of the classification models applied after the initial preprocessing using the \textit{TF-IDF} and \textit{Word2Vec} techniques. It offers a comprehensive view of their performance across multiple metrics, presenting optimized parameters alongside.

\begin{table}[H]
    \centering
    \caption{Comparison of results}
    \label{tab:comparacaotodosmodelos}
    \resizebox{\textwidth}{!}{
        \begin{tabular}{l p{4.5cm} S[table-format=2.2] S[table-format=2.2] S[table-format=2.2] S[table-format=2.2] l}
        \toprule
        Models & Best parameters &  {Accuracy (\%)} & {F1 (\%)} & {Precision (\%)} & {Recall (\%)} & Features \\
        \midrule
        Logistic Regression & C=1.0, penalty=l1, ngram\_range=1 & 95.54 & 95.49 & 96.78 & 94.24 & Tf-Idf \\
        Logistic Regression & C=0.01, penalty, l1 & 95.36 & 95.29 & 96.85 & 93.80 & Cbow \\
        Logistic Regression & C=0.01, penalty=l1 & 95.43 & 95.36 & 96.88 & 93.91 & Skipgram \\
        \midrule
        SVM & C=1, gamma=0.001, ngram\_range=1 & 93.08 & 93.19 & 91.94 & 94.49 & Tf-Idf \\
        SVM & C=100, gamma=0.001 & 55.60 & 69.25 & 53.01 & 99.94 & Cbow \\
        SVM & C=100, gamma=0.001 & 88.39 & 89.42 & 82.16 & 98.09 & Skipgram \\
        \midrule
        Random Forest & n\_estimators=20, ngram\_range=22 & 95.28 & 95.24 & 96.05 & 94.46 & Tf-Idf \\
        Random Forest & n\_estimators=100 & 95.24 & 95.24 & 95.26 & 95.24 & Cbow \\
        Random Forest & n\_estimators=100 & 95.03 & 95.00 & 95.70 & 94.32 & Skipgram \\
        \midrule
        AdaBoost & learning\_rate=1, n\_estimators=50, ngram\_range=1 & 94.63 & 94.62 & 94.74 & 94.51 & Tf-Idf \\
        AdaBoost & learning\_rate=1, n\_estimators=50 & 94.12 & 94.09 & 94.76 & 93.44 & Cbow \\
        AdaBoost & learning\_rate=1.0, n\_estimators=50 & 94.26 & 94.24 & 94.73 & 93.78 & Skipgram \\
        \midrule
        LightGBM & learning\_rate=0.3, max\_depth=8, n\_estimators=100, ngram\_range=1 & 96.17 & 96.16 & 96.48 & 95.85 & Tf-Idf \\
        LightGBM & learning\_rate=0.3, max\_depth=8, n\_estimators=100 & 95.67 & 95.65 & 96.19 & 95.13 & Cbow \\
        LightGBM & learning\_rate=0.1, max\_depth=8, n\_estimators=100 & 95.53 & 95.49 & 96.44 & 94.58 & Skipgram \\
        \bottomrule
        {\footnotesize $^{*}$Source: Prepared by the authors} & & & & & & \\
        \end{tabular}
    }
\end{table}

Our experiments' results demonstrated our approach's efficacy in detecting fake news. The LightGBM model trained on TF-IDF features achieved the highest accuracy of $96.17\%$ and an F1-Score of $96.16\%$. The other models, including logistic regression, SVM, random forest, and AdaBoost, also exhibited strong performance, with accuracy ranging from $88.39\%$ to $95.54\%$. Notably, the SVM model showed lower accuracy than other models, indicating its limitations in classifying fake news.

Furthermore, we assessed the performance of our models on recent news articles by applying the LightGBM models on the web scraping corpus, which is unseen by the model during its training. As such, for each piece of news in the set, if the probability of being true, given by the model, is $>0.5$ we mark it as true; otherwise, it is marked as fake. Table \ref{tab:comparacaobettermodels} shows the result for each gathered source, and the results show that the models can consistently differentiate between dubious and credible sources, with the CBOW and Skip-Gram Word2Vec models exhibiting greater sensitivity to dubious sources.




\begin{table}\label{tab:comparacaobettermodels}
   \centering
   \caption{Model results in current news}
       \resizebox{\linewidth}{!}{
       \begin{tabular}{lccccc}
       \toprule
       & & & \multicolumn{3}{c}{True percentages} \\ \cmidrule{4-6}
       Sources & Material & N & TF-IDF & CBOW & Skip-Gram \\ \midrule
       Jornal A & Doubtful & 26.239  & 14,04\%   & 10,72\%  & 10,69\% \\
       Jornal B & Doubtful & 2.889   & 38,80\%    & 23,23\%   & 28,35\% \\
       Jornal C & Doubtful & 15.867  & 26,41\%   & 25,25\%  &  22,60\% \\
       Brasil de fato & Credulous  & 25.412  & 80,01\%   & 80,68\%  & 80,86\% \\
       G1 & Credulous & 11.907  & 82,20\%    & 84,20\%   & 81,18\% \\
       El país & Credulous  & 16.809   & 90,78\%   & 87,54\%  & 86,66\% \\ \bottomrule
       {\footnotesize $^{*}$Source: Prepared by the authors} &  &  &  &  &  \\ 
       \end{tabular}}\label{tab:comparacaobettermodels}
\end{table}


\section{Platform for Automatic Identification of Fake News}
\label{PlatformforAutomaticIdentificationofFakeNews}
To facilitate the verification of news articles, we developed fakenewsbr.com, a user-friendly web platform. The platform allows users to paste the text of a news article for real-time verification. The verification process employs our trained models and gives users a probability score indicating the likelihood of fake news. The platform's interface is intuitive, enabling users to assess news articles' veracity quickly and conveniently.
The platform's usage is straightforward. Users access \url{http://www.fakenewsbr.com} and select the entire text of the news article to be verified. The selected text is then pasted into the "Text for Verification" field on the platform. After clicking the "Verify" button, the platform processes the text and presents the results as horizontal bars, each with a percentage. These bars represent the results of the verification process using the trained models.

\section{Conclusion and Future Work} \label{ConclusionandFutureWork}
In this paper, we presented a comprehensive study on the detection of fake news in Brazilian Portuguese, focusing on journalistic-type news articles. Our machine learning-based approach leverages natural language processing techniques and classification algorithms to achieve accurate and efficient fake news detection. The experimental results demonstrated the high performance of our models, particularly the LightGBM model trained on TF-IDF features.

The development of fakenewsbr.com further extends the practical application of our research. The platform offers a user-friendly interface for real-time verification of news articles, contributing to media literacy and informed decision-making. As part of future work, we aim to continually update our models with more recent data and explore enhancements to the platform's features and usability.

By addressing the challenge of fake news detection, we contribute to the broader efforts of combating misinformation and promoting responsible media consumption. As the landscape of information dissemination continues to evolve, our approach and platform serve as valuable tools for empowering individuals to critically assess news articles' veracity and make informed judgments.

\section*{Acknowledgments}
The authors acknowledge the support of CNPq, FAPESP, FAPEMIG, and CAPES of Brazil in funding and facilitating the research presented in this paper. 	





\section*{Appendix A: Structure of the fakenewsbr Detection Model}

This paper's fake news detection model follows a supervised analysis framework. Our input variables are constructed from the Fake.br corpus \cite{silva2020towards}, and text representations are obtained using two techniques: TF-IDF and Word2Vec. The response variable is a binary probability that indicates whether a news article is true or false. We employ various classification algorithms, including logistic regression, support vector machine, random forest, AdaBoost, and LightGBM, to perform the classification task. Model performance evaluation is based on metrics such as accuracy, precision, recall, and F1-Score.


\begin{figure}[h]
	\centering
 	  \caption{General structure of models for detecting Fake News in this work}

\tikzset{every picture/.style={line width=0.75pt}} 
\resizebox{.8\linewidth}{!}{
\begin{tikzpicture}[x=0.75pt,y=0.75pt,yscale=-1,xscale=1]

\draw   (104,144.5) -- (553,144.5) -- (553,794.5) -- (104,794.5) -- cycle ;
\draw    (100,270.5) -- (550,270.5) ;
\draw    (106,441.5) -- (556,441.5) ;
\draw    (103,538.5) -- (553,538.5) ;
\draw    (102,672) -- (552,672) ;
\draw    (160,142.5) -- (163,792.5) ;
\draw (353, 170) ellipse (28.75 and 20);
\draw (324.25,232.48) to[out=90, in=90] (381.75,232.48);
\draw    (324.25,170) -- (324.25,232.48) ;
\draw    (381.75,170) -- (381.75,232.48) ;
\draw    (353.5,253) -- (353.98,276.5) ;
\draw [shift={(354,278.5)}, rotate = 269.48] [color={rgb, 255:red, 0; green, 0; blue, 0 }  ][line width=0.75]    (10.93,-3.29) .. controls (6.95,-1.4) and (3.31,-0.3) .. (0,0) .. controls (3.31,0.3) and (6.95,1.4) .. (10.93,3.29)   ;
\draw   (177,309.7) .. controls (177,293.3) and (190.3,280) .. (206.7,280) -- (505.3,280) .. controls (521.7,280) and (535,293.3) .. (535,309.7) -- (535,398.8) .. controls (535,415.2) and (521.7,428.5) .. (505.3,428.5) -- (206.7,428.5) .. controls (190.3,428.5) and (177,415.2) .. (177,398.8) -- cycle ;
\draw   (178,472.9) .. controls (178,464.95) and (184.45,458.5) .. (192.4,458.5) -- (521.6,458.5) .. controls (529.55,458.5) and (536,464.95) .. (536,472.9) -- (536,516.1) .. controls (536,524.05) and (529.55,530.5) .. (521.6,530.5) -- (192.4,530.5) .. controls (184.45,530.5) and (178,524.05) .. (178,516.1) -- cycle ;
\draw   (179,572.1) .. controls (179,559.89) and (188.89,550) .. (201.1,550) -- (514.9,550) .. controls (527.11,550) and (537,559.89) .. (537,572.1) -- (537,638.4) .. controls (537,650.61) and (527.11,660.5) .. (514.9,660.5) -- (201.1,660.5) .. controls (188.89,660.5) and (179,650.61) .. (179,638.4) -- cycle ;
\draw   (181,703.3) .. controls (181,691.54) and (190.54,682) .. (202.3,682) -- (517.7,682) .. controls (529.46,682) and (539,691.54) .. (539,703.3) -- (539,767.2) .. controls (539,778.96) and (529.46,788.5) .. (517.7,788.5) -- (202.3,788.5) .. controls (190.54,788.5) and (181,778.96) .. (181,767.2) -- cycle ;
\draw    (357.5,430) -- (357.96,453.5) ;
\draw [shift={(358,455.5)}, rotate = 268.88] [color={rgb, 255:red, 0; green, 0; blue, 0 }  ][line width=0.75]    (10.93,-3.29) .. controls (6.95,-1.4) and (3.31,-0.3) .. (0,0) .. controls (3.31,0.3) and (6.95,1.4) .. (10.93,3.29)   ;
\draw    (359.5,528) -- (359.94,543.5) ;
\draw [shift={(360,545.5)}, rotate = 268.36] [color={rgb, 255:red, 0; green, 0; blue, 0 }  ][line width=0.75]    (10.93,-3.29) .. controls (6.95,-1.4) and (3.31,-0.3) .. (0,0) .. controls (3.31,0.3) and (6.95,1.4) .. (10.93,3.29)   ;
\draw    (359.5,658) -- (359.95,677.5) ;
\draw [shift={(360,679.5)}, rotate = 268.67] [color={rgb, 255:red, 0; green, 0; blue, 0 }  ][line width=0.75]    (10.93,-3.29) .. controls (6.95,-1.4) and (3.31,-0.3) .. (0,0) .. controls (3.31,0.3) and (6.95,1.4) .. (10.93,3.29)   ;


\draw (327, 192) node [anchor=north west][inner sep=0.75pt]   [align=left] 
{\begin{minipage}[lt]{37.3pt}\setlength\topsep{0pt}
\begin{center}
Fake.br\\Corpus
\end{center}

\end{minipage}};
\draw (200.3,283) node [anchor=north west][inner sep=0.75pt]   [align=left] {Removal of special characters};
\draw (201,303) node [anchor=north west][inner sep=0.75pt]   [align=left] {Removing links and html};
\draw (202,323) node [anchor=north west][inner sep=0.75pt]   [align=left] {Accent removal};
\draw (201,342) node [anchor=north west][inner sep=0.75pt]   [align=left] {Stop word removal};
\draw (202,363) node [anchor=north west][inner sep=0.75pt]   [align=left] {Removing scores};
\draw (202,382) node [anchor=north west][inner sep=0.75pt]   [align=left] {Stemming or Bigram creation};
\draw (203,403) node [anchor=north west][inner sep=0.75pt]   [align=left] {Vectorization of news in sentences};
\draw (204,487) node [anchor=north west][inner sep=0.75pt]   [align=left] {TF-IDF, Skip-Gram, CBOW};
\draw (206,559) node [anchor=north west][inner sep=0.75pt]   [align=left] {Logistic Regression};
\draw (206,580) node [anchor=north west][inner sep=0.75pt]   [align=left] {SVM};
\draw (207,599) node [anchor=north west][inner sep=0.75pt]   [align=left] {Random Forest};
\draw (208,618) node [anchor=north west][inner sep=0.75pt]   [align=left] {AdaBoost};
\draw (208,636) node [anchor=north west][inner sep=0.75pt]   [align=left] {LightGBM};
\draw (208,694) node [anchor=north west][inner sep=0.75pt]   [align=left] {Accuracy};
\draw (209,714) node [anchor=north west][inner sep=0.75pt]   [align=left] {F1 Score};
\draw (209,735) node [anchor=north west][inner sep=0.75pt]   [align=left] {Precision};
\draw (209,755) node [anchor=north west][inner sep=0.75pt]   [align=left] {Recall};
\draw (127.38,771.55) node [anchor=north west][inner sep=0.75pt]  [rotate=-269.77] [align=left] {Evaluation};
\draw (127.94,633.54) node [anchor=north west][inner sep=0.75pt]  [rotate=-269.77] [align=left] {Models};
\draw (127.94,531.44) node [anchor=north west][inner sep=0.75pt]  [rotate=-270.4] [align=left] {\begin{minipage}[lt]{58.84pt}\setlength\topsep{0pt}
Feature set
\end{minipage}};
\draw (126.85,419.54) node [anchor=north west][inner sep=0.75pt]  [rotate=-269.85] [align=left] {Pre-Processing};
\draw (126.93,232.61) node [anchor=north west][inner sep=0.75pt]  [rotate=-269.31] [align=left] {Data};

\end{tikzpicture}
}
\\
	 {\footnotesize Source: Prepared by the authors} 
	\label{fig:figura-lgbm_w1v}
\end{figure}


\begin{figure}[h]
	\centering
        \caption{Structure of training the Word2Vec models on the web scrapped data}

\tikzset{every picture/.style={line width=0.75pt}} 
\resizebox{.8\linewidth}{!}{
\begin{tikzpicture}[x=0.75pt,y=0.75pt,yscale=-1,xscale=1]

\draw   (104,144.5) -- (553,144.5) -- (553,673.5) -- (104,673.5) -- cycle ;
\draw    (100,270.5) -- (550,270.5) ;
\draw    (106,441.5) -- (556,441.5) ;
\draw    (103,538.5) -- (553,538.5) ;
\draw    (160,142.5) -- (163,673.5) ;
\draw   (303,60) -- (400,60) -- (400,129.5) -- (303,129.5) -- cycle ;
\draw    (351.5,131) -- (351.5,149) ;
\draw [shift={(351.5,149)}, rotate = 270] [color={rgb, 255:red, 0; green, 0; blue, 0 }  ][line width=0.75]    (10.93,-3.29) .. controls (6.95,-1.4) and (3.31,-0.3) .. (0,0) .. controls (3.31,0.3) and (6.95,1.4) .. (10.93,3.29)   ;
\draw (353, 170) ellipse (28.75 and 20);
\draw (324.25,232.48) to[out=90, in=90] (381.75,232.48);
\draw    (324.25,170) -- (324.25,232.48) ;
\draw    (381.75,170) -- (381.75,232.48) ;
\draw    (353.5,253) -- (353.98,276.5) ;
\draw [shift={(354,278.5)}, rotate = 269.48] [color={rgb, 255:red, 0; green, 0; blue, 0 }  ][line width=0.75]    (10.93,-3.29) .. controls (6.95,-1.4) and (3.31,-0.3) .. (0,0) .. controls (3.31,0.3) and (6.95,1.4) .. (10.93,3.29)   ;
\draw   (177,309.7) .. controls (177,293.3) and (190.3,280) .. (206.7,280) -- (505.3,280) .. controls (521.7,280) and (535,293.3) .. (535,309.7) -- (535,398.8) .. controls (535,415.2) and (521.7,428.5) .. (505.3,428.5) -- (206.7,428.5) .. controls (190.3,428.5) and (177,415.2) .. (177,398.8) -- cycle ;
\draw    (353.5,430) -- (353.5,455) ;
\draw [shift={(354,456)}, rotate = 269.48] [color={rgb, 255:red, 0; green, 0; blue, 0 }  ][line width=0.75]    (10.93,-3.29) .. controls (6.95,-1.4) and (3.31,-0.3) .. (0,0) .. controls (3.31,0.3) and (6.95,1.4) .. (10.93,3.29)   ;
\draw   (178,472.9) .. controls (178,464.95) and (184.45,458.5) .. (192.4,458.5) -- (521.6,458.5) .. controls (529.55,458.5) and (536,464.95) .. (536,472.9) -- (536,516.1) .. controls (536,524.05) and (529.55,530.5) .. (521.6,530.5) -- (192.4,530.5) .. controls (184.45,530.5) and (178,524.05) .. (178,516.1) -- cycle ;
\draw    (353.5,532) -- (353.98,550) ;
\draw [shift={(354,551)}, rotate = 269.48] [color={rgb, 255:red, 0; green, 0; blue, 0 }  ][line width=0.75]    (10.93,-3.29) .. controls (6.95,-1.4) and (3.31,-0.3) .. (0,0) .. controls (3.31,0.3) and (6.95,1.4) .. (10.93,3.29)   ;
\draw   (179,572.1) .. controls (179,559.89) and (188.89,550) .. (201.1,550) -- (514.9,550) .. controls (527.11,550) and (537,559.89) .. (537,572.1) -- (537,638.4) .. controls (537,650.61) and (527.11,660.5) .. (514.9,660.5) -- (201.1,660.5) .. controls (188.89,660.5) and (179,650.61) .. (179,638.4) -- cycle ;

\draw (305,80.5) node [anchor=north west][inner sep=0.75pt]   [align=left] {\begin{minipage}[lt]{66.61pt}\setlength\topsep{0pt}
\begin{center}
Web Scraping\\News
\end{center}

\end{minipage}};
\draw (335, 195) node [anchor=north west][inner sep=0.75pt]   [align=left] {News\\Bank};
\draw (200.3,283) node [anchor=north west][inner sep=0.75pt]   [align=left] {Removal of special characters};
\draw (201,303) node [anchor=north west][inner sep=0.75pt]   [align=left] {Removing links and html};
\draw (202,323) node [anchor=north west][inner sep=0.75pt]   [align=left] {Accent removal};
\draw (201,342) node [anchor=north west][inner sep=0.75pt]   [align=left] {Stop word removal};
\draw (202,363) node [anchor=north west][inner sep=0.75pt]   [align=left] {Removing scores};
\draw (202,382) node [anchor=north west][inner sep=0.75pt]   [align=left] {Bigram creation};
\draw (203,403) node [anchor=north west][inner sep=0.75pt]   [align=left] {Vectorization of news in sentences};
\draw (204,477) node [anchor=north west][inner sep=0.75pt]   [align=left] {Word2Vec (CBOW)};
\draw (203,497) node [anchor=north west][inner sep=0.75pt]   [align=left] {Word2Vec (Skip-Gram)};
\draw (206,559) node [anchor=north west][inner sep=0.75pt]   [align=left] {min\_count = 20};
\draw (206,580) node [anchor=north west][inner sep=0.75pt]   [align=left] {window = 2};
\draw (206,599) node [anchor=north west][inner sep=0.75pt]   [align=left] {alpha = 0.03};
\draw (206,618) node [anchor=north west][inner sep=0.75pt]   [align=left] {min\_alpha = 0.0007};
\draw (127.94,510.44) node [anchor=north west][inner sep=0.75pt]  [rotate=-270.4] [align=left] {Models};
\draw (127.94,633.54) node [anchor=north west][inner sep=0.75pt]  [rotate=-269.77] [align=left] {Parameters};


\draw (126.85,419.54) node [anchor=north west][inner sep=0.75pt]  [rotate=-269.85] [align=left] {Pre-Processing};
\draw (126.93,232.61) node [anchor=north west][inner sep=0.75pt]  [rotate=-269.31] [align=left] {Data};

\end{tikzpicture}
}
\\
	 {\footnotesize Source: Prepared by the authors} 
	\label{fig:figura-lgbm_w2v}
\end{figure}


\section*{Appendix B: Platform for Automatic Identification of Fake News}
The \url{http://www.fakenewsbr.com} platform provides a user-friendly solution for automatically identifying fake news. Users can easily verify the veracity of news articles by pasting the article's text into the platform. After clicking the "Verify" button, the platform processes the text using trained classification models and presents the results as horizontal bars. These bars indicate the probability of the fake news article, as determined by the models. The platform's intuitive interface and real-time analysis enable users to make informed decisions about the credibility of online news articles.

\begin{figure}[H]
  \centering
  \includegraphics[width=0.9\linewidth]{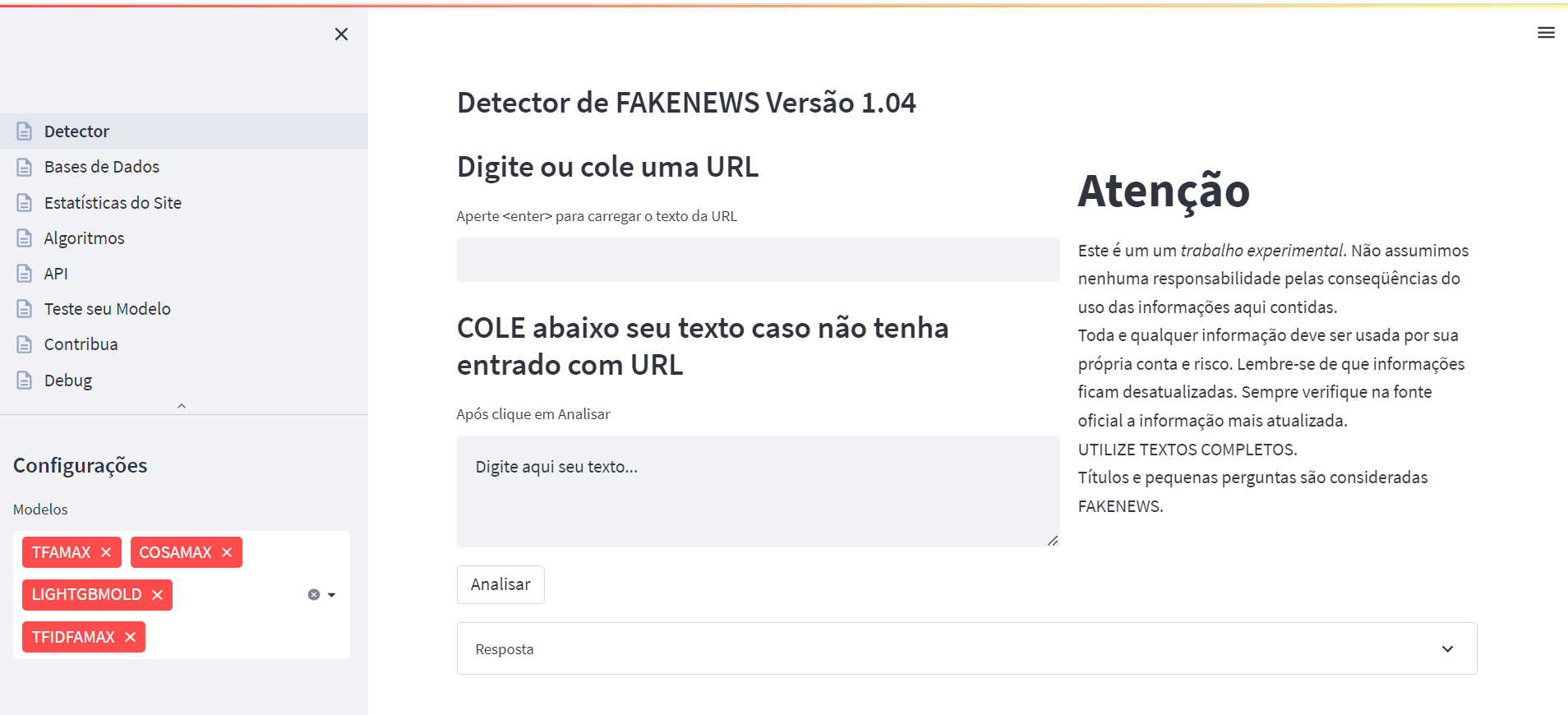}
  \caption{Web Platform for Automatic Fake News Detection.}
  \label{fig:Foto-Plataforma-Fake-News-1}
\end{figure}

\begin{figure}[H]
  \centering
  \includegraphics[width=0.9\linewidth]{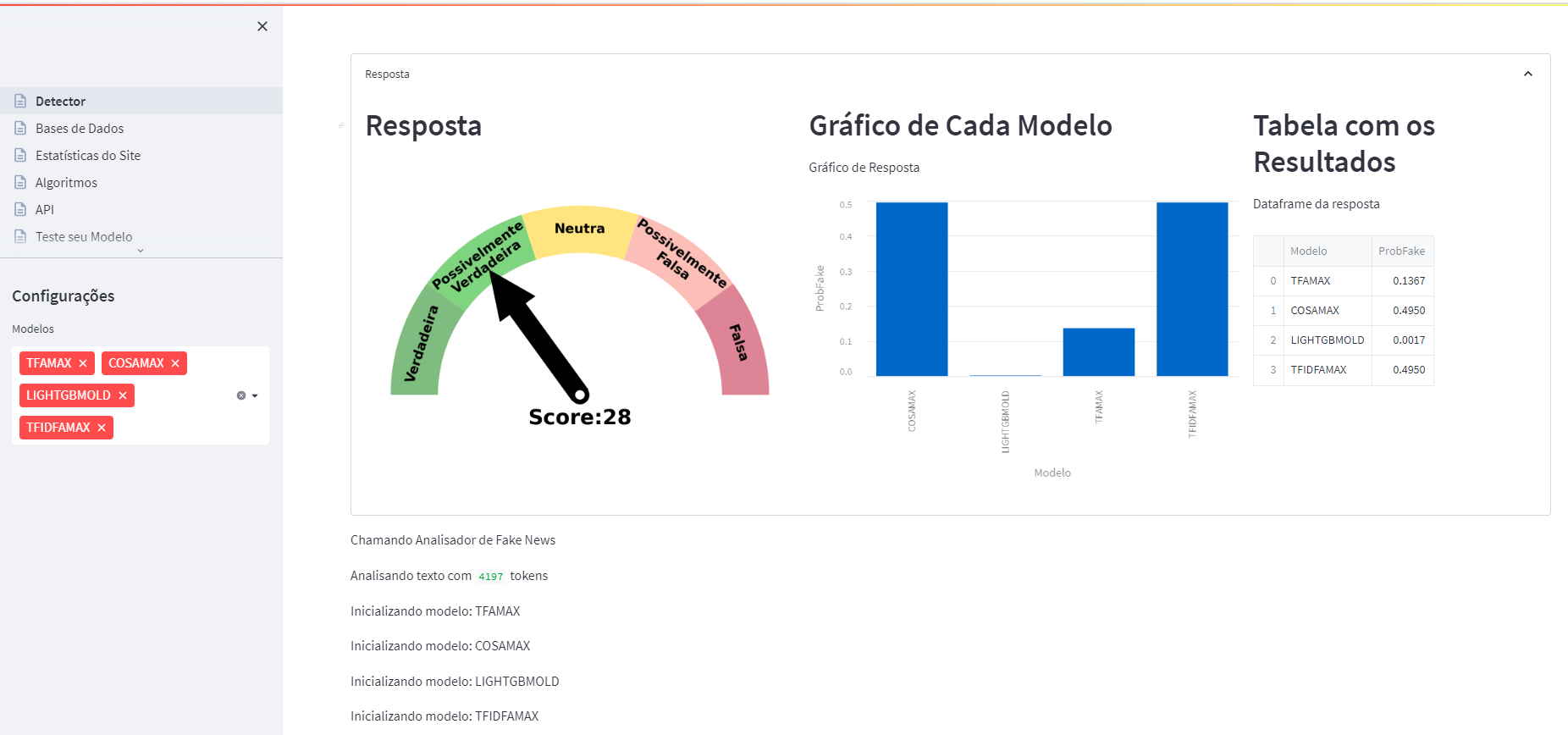}
  \caption{Results provided by the fake news verification platform.}
  \label{fig:Foto-Plataforma-Fake-News-4}
\end{figure}

The following is a step-by-step description of properly using the Journalistic News Verification Platform.
\begin{itemize}
  \item \textbf{Step 1: Access the Platform} \\
  To begin, access the web platform at \url{http://www.fakenewsbr.com}.

  \item \textbf{Step 2: User-Friendly Interface} \\
  Upon reaching the initial screen (Figure \ref{fig:Foto-Plataforma-Fake-News-1}), you'll find a user-friendly interface for easy use.

  \item \textbf{Step 3: Select News Text} \\
  Select the entire text of the news article one want to verify.

  \item \textbf{Step 4: Copy the Text} \\
  Copy the selected text to the one clipboard.

  \item \textbf{Step 5: Paste the Text} \\
  Paste the copied text into the "Text for Verification" field, located in the central region of the platform (Figure \ref{fig:Foto-Plataforma-Fake-News-1}).

  \item \textbf{Step 6: Verification} \\
  Click the "Verify" button below the text input box.

  \item \textbf{Step 7: Wait for Results} \\
  Wait a few seconds for the results to appear on the screen (Figure \ref{fig:Foto-Plataforma-Fake-News-4}).

  \item \textbf{Step 8: Interpret Results} \\
  The results are presented as horizontal green bars, each with a percentage at the beginning. These bars represent the outcomes of four statistical models that have been trained and tested to verify the news's accuracy. Additionally, the last bar has a weighted average of these results. 
  
 Now, users can make informed decisions about the veracity of the news they have examined.
\end{itemize}

\begin{figure}[H]
  \centering
  \includegraphics[width=0.9\linewidth]{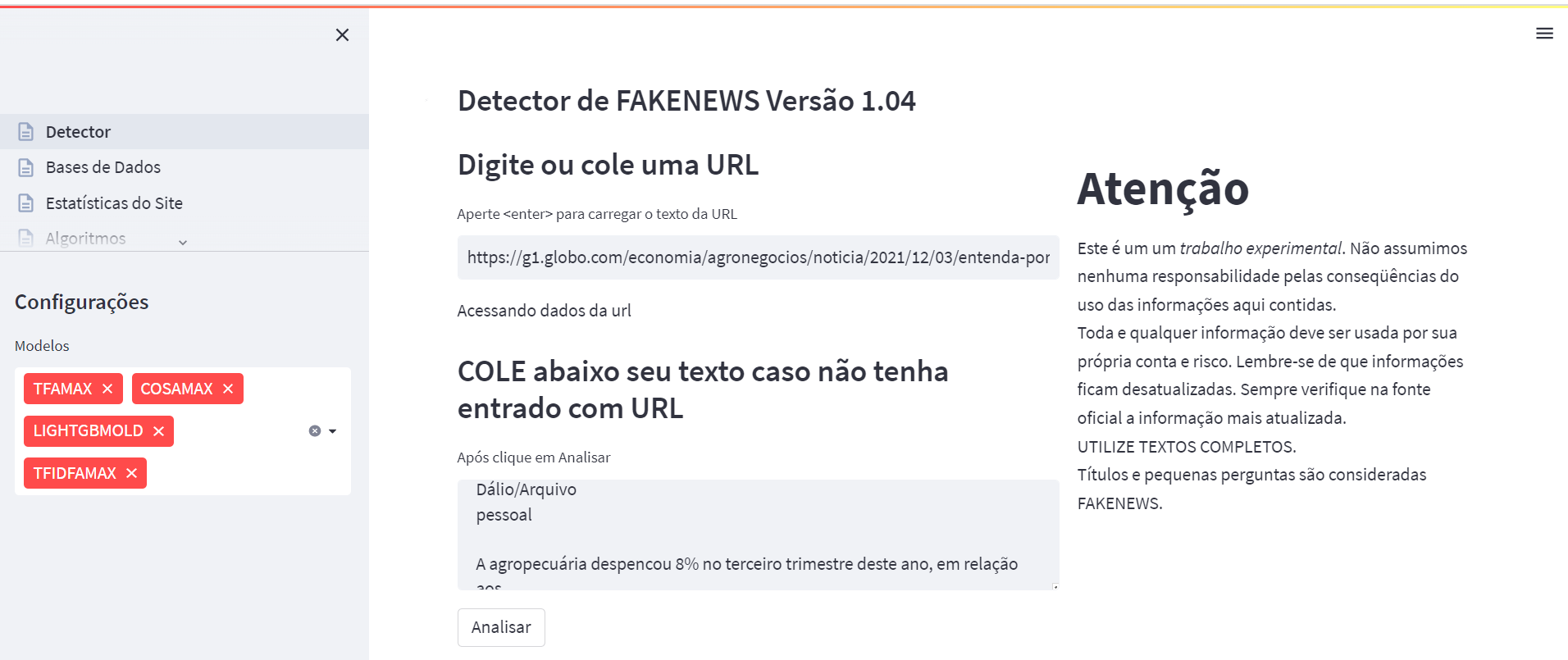}
  \caption{Pasting the chosen news in the verification platform.}
  \label{fig:Foto-Plataforma-Fake-News-3}
\end{figure}

\bibliographystyle{tfs}
\bibliography{interacttfssample}

\end{document}